\documentclass{article}
\usepackage{ijcai24}

\usepackage{times}
\usepackage{soul}
\usepackage{url}
\usepackage[hidelinks]{hyperref}
\usepackage[utf8]{inputenc}
\usepackage[small]{caption}
\usepackage{graphicx}
\usepackage{amsmath}
\usepackage{amsthm}
\usepackage{booktabs}
\usepackage{algorithm}
\usepackage{algorithmic}

\usepackage{multirow}
\usepackage{amssymb}
\usepackage{pifont}
\usepackage{amsfonts}

\usepackage[switch]{lineno}

\title{CBVS: A Large-Scale Chinese Image-Text Benchmark for Real-World Short Video Search Scenarios}

\author{
Xiangshuo Qiao$^{1*}$
\and
Xianxin Li$^{1}$\footnote{Co-first authors}\and
Xiaozhe Qu$^1$\and
Jie Zhang$^{1}$\footnote{Corresponding author}\and
Yang Liu$^1$\and
Yu Luo$^1$\and
Cihang Jin$^1$\and
Jin Ma$^2$
\affiliations
$^1$Tencent PCG\\
$^2$University of Science and Technology of China
\emails
\{xsqiao, ceceliali, xiaozhequ, jeyzzhang, jelmeliu, yamiluo, alexajin\}@tencent.com
\{majin01\}@mail.ustc.edu.cn
}

\begin{document}

\maketitle

\begin{abstract}
Vision-Language Models pre-trained on large-scale image-text datasets have shown superior performance in downstream tasks such as image retrieval. Most of the images for pre-training are presented in the form of open domain common-sense visual elements. Differently, video covers in short video search scenarios are presented as user-originated contents that provide important visual summaries of videos. In addition, a portion of the video covers come with manually designed cover texts that provide semantic complements. In order to fill in the gaps in short video cover data, we establish the first large-scale cover-text benchmark for Chinese short video search scenarios. Specifically, we release two large-scale datasets CBVS-5M/10M to provide short video covers, and the manual fine-labeling dataset CBVS-20K to provide real user queries, which serves as an image-text benchmark test in the Chinese short video search field. To integrate the semantics of cover text in the case of modality missing, we propose UniCLIP where cover texts play a guiding role during training, however are not relied upon by inference. Extensive evaluation on CBVS-20K demonstrates the excellent performance of our proposal. UniCLIP has been deployed to Tencent's online video search systems with hundreds of millions of visits and achieved significant gains. The dataset and code are available at \url{https://github.com/QQBrowserVideoSearch/CBVS-UniCLIP}.
\end{abstract}

\section{Introduction}

\begin{figure}
\begin{center}
\includegraphics[width=\linewidth]{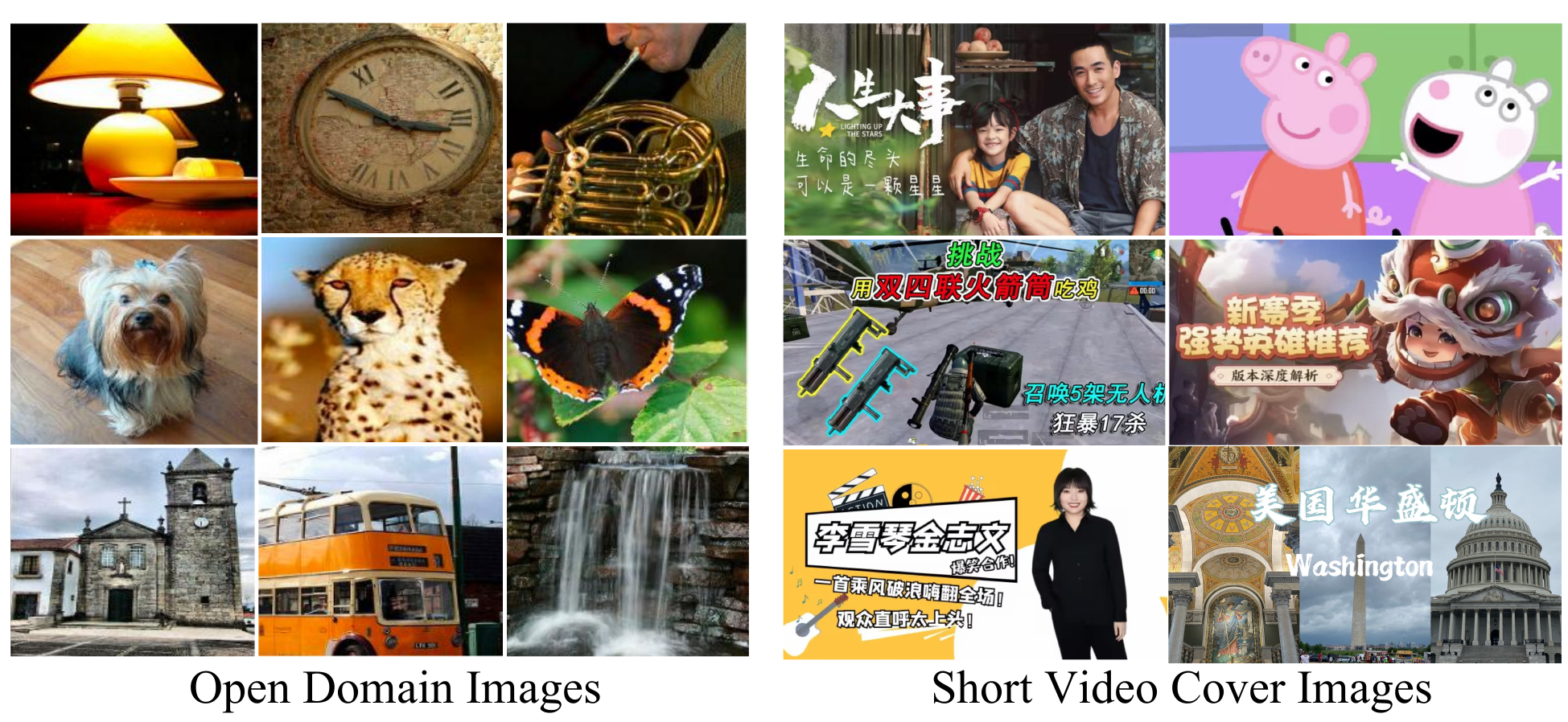}
\end{center}
   \caption{Morphological differences between open domain images and short video cover images.}
\label{fig:gap}
\end{figure}

\begin{figure*}
\begin{center}
\includegraphics[width=\linewidth]{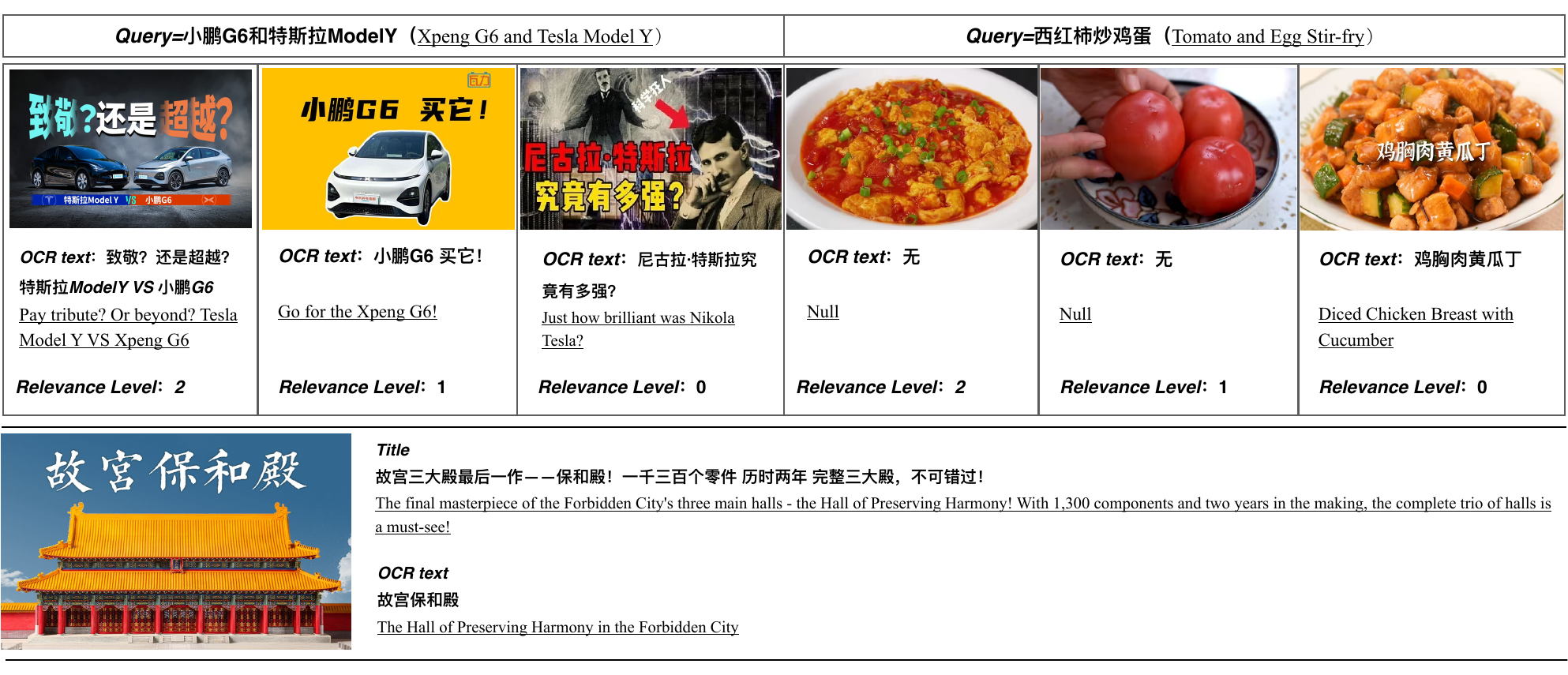}
\end{center}
   \caption{Top: Presentation of CBVS-20K data. Bottom: Presentation of CBVS-5M/10M data.}
\label{fig:case}
\end{figure*}

CLIP \cite{radford2021learning} demonstrates the promise of performing contrastive learning pre-training on large-scale image-text data from the web with a data size of 400 million. In this work, visual base models represented by ViT \cite{dosovitskiy2020image} are aligned with textual base models represented by Bert \cite{vaswani2017attention,devlin2018bert} by learning on large-scale unsupervised data. These base models can be transferred to downstream tasks such as image search \cite{hendriksen2022extending} via natural language prompts \cite{zhou2022learning}.

In the field of Chinese multi-modal representation learning, previous work \cite{yang2022chinese,xie2023ccmb,gu2022wukong} supplements high-quality Chinese image-text datasets and successfully pre-trains Chinese visual language models. Most of the data are open-domain images collected from the web or multiplexed from publicly available English datasets. These images are captured by a camera and presented in the form of common-sense visual elements, including animals, buildings, activities, etc. with corresponding descriptive text.

With the rise of short videos, video search has become a popular topic \cite{spolaor2020systematic,wray2021semantic}. Previous work \cite{zhang2022create,nie2022search,xu2023youku} create large-scale datasets for the Chinese short-video search domain and provide publicly available video frames or video features to support content-based search. However, cover-based search remains to be investigated. During the creation of short videos, creators craft video covers for short videos with the aim of attracting the interest of the most relevant viewers. Therefore, in short video search scenarios, short video covers provide direct overviews and serve as crucial visual features of the videos. Besides, cover-based search has efficiency advantages over content-based search.

However, there are remarkable morphological differences between short video cover images and open domain images. As shown in Fig. \ref{fig:gap}, compared to open-domain visual elements, short video covers, as user-originated content, are mostly artificial combinations of various visual elements and may undergo post-processing such as cropping and splicing. On the other hand, many creators craft cover texts for video covers to complement or emphasize the semantic information of the video. This is a feature that open domain images do not share. Therefore, short video cover images represent a different form of data from open domain images, and the availability of large-scale cover dataset is crucial. However, available large-scale cover datasets are lacking.

In this work, we release a large-scale Chinese image-text Benchmark for short Video Search scenarios (CBVS) to fill the gap of data in real Chinese video search scenarios. CBVS is designed in three versions: the manually fine-labeled CBVS-20K and the large-scale unsupervised CBVS-5M/10M. Fig. \ref{fig:case} shows their data examples. Specifically, CBVS-20K contains 20K high-quality $<$user query-video cover$>$ pairs, which serves as an image-text benchmark test in the field of Chinese short video search. Well-trained human experts annotate the relevance of each user query to the video cover and at least two cross-validations are performed. In addition, Optical Character Recognition (OCR) texts of cover images are provided after machine extraction and human correction. Due to the constraints of user privacy and platform rules, the large-scale CBVS-5M/10M contains about 5M/10M $<$video title-video cover$>$ pairs, where the text is provided in the form of video titles and OCR texts. These data are available for visual language models to learn modal alignment in pre-training or fine-tuning tasks. The CBVS dataset includes 32 categories such as Film and animation, Character, Education, Game, Commodity, etc. to avoid data distribution bias. Tab. \ref{tab:cmp_version} shows a detailed comparison of various versions.

In short video search scenarios, cover texts complement the semantics of cover images. On one hand, CLIP lacks the ability to fuse multi-semantic signals on the visual side. On the other hand, not all cover images come with cover texts, so the modality missing problem needs to be considered. In order to effectively integrate the semantics of cover images with cover texts, we propose UniCLIP inspired by the work of OCR-free \cite{kim2022ocr,davis2022end}. Cover texts signals are unified to guide image-text contrastive learning in an presence-guided and semantic-guided manner. It is worth emphasizing that the inference process does not depend on any module related to OCR and the model is immune to the problem of missing cover text modalities. Extensive experimental evaluations demonstrate the effectiveness of our proposal.

Our contributions can be summarized as follows:
\begin{itemize}
    \item In order to fill in the lack of cover data for short video search scenarios, we release the largest Chinese cover image-text dataset with video title texts and cover texts.
    \item We build a manual fine-labeling image-text benchmark test for Chinese short video search scenarios, containing real user queries from browser logs.
    \item We propose UniCLIP, which introduces an image classification task and an image-text matching task to guide image-text contrastive learning training. UniCLIP imposes no additional inference cost and training is immune to the modality missing problem.

\end{itemize}

\begin{table*}
\begin{center}
    \caption{Statistics of Chinese image-text/video-text datasets.}
    \label{tab:cmp_dataset}
    \begin{tabular}{c|c|c|c|c|c|c}
    \toprule
    Dataset&$\#$Vision&$\#$Text&Source&Vision type&Text type&Availability\\
    \midrule
    \multicolumn{7}{c}{Chinese Video-Text Datasets}\\
    \midrule
    VATEX&41,269&825,380&Kinetics-600&Video&Caption&\checkmark\\
    BFVD&43,166&43,166&E-Commerce&Feature&Title&\checkmark\\
    FFVD&32,763&32,763&E-Commerce&Feature&Title&\checkmark\\
    CREATE-210K&216,303&268,593&Open Websites&Video&Title, Caption&\ding{55}\\
    CREATE-3M&3,000,000&3,000,000&Open Websites&Video&Title&\ding{55}\\
    CREATE-10M&10,000,000&10,000,000&Open Websites&Video&Title&\ding{55}\\
    Kwai-SVC&222,077&143,569&Video Search&Feature&Title, OCR, ASR&\ding{55}\\
    Kwai-SVC-11M&11,075,084&3,931,879&Video Search&Video&Title, ASR&\ding{55}\\
    CNVid-3.5M&3,508,120&3,508,120&Open Websites&Video&ASR, Title&\checkmark\\
    ALIVOL-10M&10,300,000&11,000,000&E-Commerce&Video, Image&Title, Abstract&\ding{55}\\
    Youku-mPLUG&10,000,000&10,000,000&Open Websites&Video&Title&\checkmark\\
    \midrule
    \multicolumn{7}{c}{Chinese Image-Text Datasets}\\
    \midrule
    Wukong&101,483,885&101,483,885&Open Websites&Image&Caption&\checkmark\\
    Wukong-Test&33,365&33,365&Open Websites&Image&Caption&\checkmark\\
    Product1M&1,182,083&1,182,083&E-Commerce&Image&Caption&\checkmark\\
    M6-Corpus&60,500,000&60,500,000&Open Websites&Image&Caption&\ding{55}\\
    ZERO-Corpus&250,000,000&750,000,000&Image Search&Image&Title, Content, Query&\checkmark\\
    R2D2-ICR&200,000&200,000&Image Search&Image&Caption&\checkmark\\
    R2D2-IQR&200,000&200,000&Image Search&Image&Query&\checkmark\\
    CBVS-20K&20,001&20,001&Video Search&Cover Image&OCR, Query&\checkmark\\
    CBVS-5M&4,767,435&4,767,435&Video Search&Cover Image&OCR, Title&\checkmark\\
    CBVS-10M&10,075,989&10,075,989&Video Search&Cover Image&OCR, Title&\checkmark\\
    \bottomrule
    \end{tabular}
\end{center}
\end{table*}

\section{Related Work}

\subsection{Chinese Video/Image-text benchmark}

Compared to English multi-modal pre-training, the Chinese community is lagging behind. \cite{yang2022chinese} introduces translated versions of the English multi-modal datasets \cite{chen2015microsoft,krishna2017visual} to support Chinese multi-modal pre-training. \cite{gu2022wukong} releases a large-scale Chinese dataset Wukong containing 100 million image-text pairs collected from the web to bridge the language gap. \cite{xie2023ccmb} further establishes large-scale Chinese cross-modal benchmarks by releasing two pre-training datasets and five fine-tuning datasets. Besides, Product1M \cite{zhan2021product1m} provides additions to the e-commerce domain. \cite{zhang2022create,nie2022search,xu2023youku} supplements the Chinese video-text data, and visual modalities are provided in the form of video frames. However, large-scale video cover data is scarce.

\begin{table}
\begin{center}
    \caption{Comparison of CBVS versions.}
    \label{tab:cmp_version}
    \begin{tabular}{c|c|c|c|c}
    \toprule
    Version&\#Pair&Purp.&Ann.&Text type\\
    \midrule
    CBVS-20K&20,001&Test&\checkmark&Query, OCR\\
    CBVS-5M&4,767,435&Train&\ding{55}&Title, OCR\\
    CBVS-10M&10,075,989&Train&\ding{55}&Title, OCR\\
    \bottomrule
    \end{tabular}
\end{center}
\end{table}

\subsection{Image-text Matching}

The image-text matching task aims to measure the semantic similarity of different modalities in the same embedding space\cite{abdullah2021image}. Existing implementations fall into two categories: the first is embedding-based, i.e., encoding global representations of vision and text separately, and then performing similarity computation \cite{chen2021learning,faghri2017vse++,qu2020context}. The second is score-based, i.e., performing cross-modal interactions locally and calculating cumulative scores \cite{chen2020imram,diao2021similarity,liu2020graph}. Due to the advantages of performance and efficiency, embedding-based methods have attracted the attention of researchers \cite{fu2023learning}. In particular, CLIP \cite{radford2021learning} provides new ideas for multi-modal representation learning. A series of studies following CLIP have been applied to downstream tasks including image text matching. For example, GCLIP\cite{li2022grounded}, CLIP-Adapter\cite{zhang2021tip} and SLIP\cite{mu2022slip} raise the upper performance limit of CLIP, AltCLIP\cite{chen2022altclip}, CN-CLIP\cite{yang2022chinese}, and TaiyiCLIP\cite{zhang2022fengshenbang} expand the CLIP language domain.

% https://blog.csdn.net/xovee/article/details/122179352

\section{CBVS Dataset}

\begin{figure*}
\begin{center}
\includegraphics[width=\linewidth]{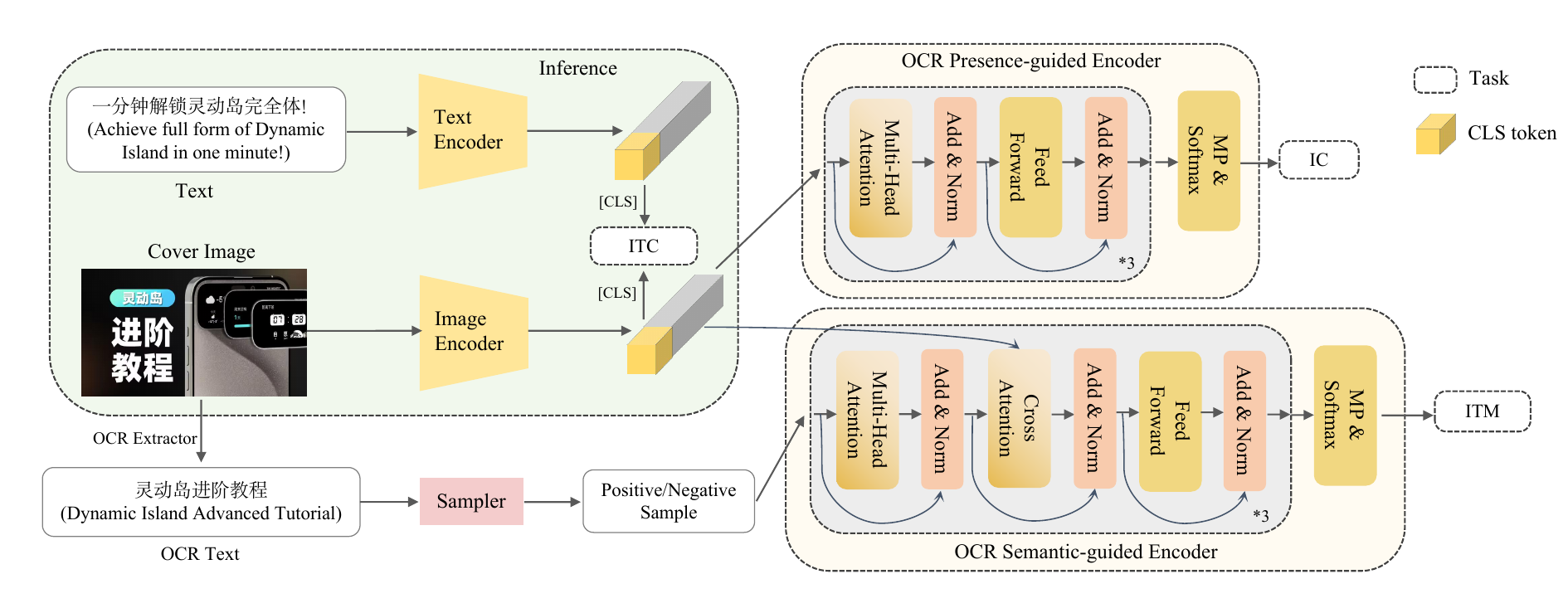}
\end{center}
   \caption{Model structure of UniCLIP. When the model performs inference, only the green area works.}
\label{fig:model}
\end{figure*}

\subsection{Comparison}

A comparison with other Chinese image-text/video-text datasets is shown in Tab. \ref{tab:cmp_dataset}. On the one hand, CBVS provides cover images, user queries, and is larger in size compared to publicly available video datasets such as VATEX \cite{wang2019vatex}, BFVD/FFVD \cite{zhang2020poet}, and CNVid-3.5M \cite{gan2023cnvid}. CREATE \cite{zhang2022create}, Kwai-SVC \cite{nie2022search}, ALIVOL-10M \cite{lei2021understanding} with similar scale are access restricted. On the other hand, compared to image-text datasets such as Wukong \cite{gu2022wukong}, Product1M \cite{zhan2021product1m}, M6-Corpus \cite{lin2021m6}, ZERO-Corpus/R2D2 \cite{xie2023ccmb}, etc., the biggest advantage of CBVS lies in the uniqueness of the video cover image and the cover text specific to the cover image. To the best of our knowledge, CBVS is the largest publicly available Chinese image-text dataset providing cover images.

\subsection{Data Collection}

In order to provide real video search data, we capture the user query logs of QQ Browser and divide user queries into two parts. We retrieve more than 8M videos from the Chinese video website BiliBili\footnote{https://www.bilibili.com} through the first part of user queries. To avoid data distribution bias due to a single platform, we retrieve more than 5M videos from Tencent Video\footnote{https://v.qq.com} as a supplement in the same way, and finally obtain more than 13M $<$cover-title$>$ pairs as a data source for CBVS-5M/10M. Besides, we manually select more than 2K high-quality user queries from the second part, and collect 20K high-quality $<$user query-cover image$>$ pairs in the same way, as the data source of CBVS-20K. The number of cover images under each user query is controlled in $[5,30)$. 

We design the data cleaning program from two aspects: data quality and image-text relevance. First, we filter the video covers with low resolution and scale disproportion, and eliminate the dead links. After that, we score the relevance of video covers and titles in 13M data with the open-source Chinese image-text model QA-CLIP\footnote{https://github.com/TencentARC-QQ/QA-CLIP}, and filter the trailing 3M data to obtain CBVS-10M. We randomly sample 1K data from 13M and 10M data, respectively, and human experts evaluate whether the video cover is relevant to the title. The evaluation conclusions show that the relevance is improved from 75.6\% to 93.0\% after data cleaning. Finally, CBVS-5M is obtained by sampling in CBVS-10M.

\subsection{Data Annotation}

The data annotation of CBVS-20K is performed by trained experts in the field of video search in a two-stage, cross-validated approach. First, they annotate whether the user query reveals a clear need for video, and filter out query terms related to pornography and violence, and finally take the queries with a need for video as candidates for annotation in the second stage.  After that, the annotators mark the degree of relevance for each $<$user query-cover image$>$ pair, which is categorized into three grades: strongly relevant, weakly relevant, and irrelevant. The semantics of cover images and cover texts are required to be considered together. Meanwhile, the annotators correct the OCR text extracted by the machine.

We exclude the controversial data, and end up with 20,001 image-text pairs consisting of 2,486 unique queries and 19,648 unique images. The percentage of strongly relevant, weakly relevant and irrelevant data is 29.74\%, 30.80\% and 39.46\% respectively. The average length of user queries is 7.0 and the average length of OCR texts is 14.5. 33.41\% of cover images come with OCR texts. Detailed data distribution is provided in the supplementary material.

\section{Methodology}

\subsection{Image-Text Contrastive Learning}

We follow CLIP \cite{radford2021learning} to co-train the image encoder with the text encoder, taking InfoNCE Loss \cite{oord2018representation} as the Image-Text Contrastive (ITC) loss $L_{ITC}$, as shown in Fig. \ref{fig:model}. Specifically, we maximize the similarity scores of the matched image and text embeddings in terms of batch. We adopt ViT \cite{dosovitskiy2020image} and RoBERTa \cite{liu2019roberta} as the visual and textual skeletons, respectively, and introduce the weight initialization of QA-CLIP.

In particular, to bridge the gap between the data morphology of the title and the user query, we employ a Chinese word-splitting component, Lexical Analysis of Chinese (LAC)\footnote{https://github.com/baidu/lac}. In order to simulate the morphological distribution of the user query, the result of the lexical segmentation is composed into a string using spaces as the splice character. For the case of failed word splitting, the original title is employed. This setting takes effect for all fine-tuning tasks unless otherwise specified.

\subsection{Presence-guided Encoder}

Video cover images differ from open domain images in that they partially carry cover texts. One option is to outsource the cover text understanding task to an external OCR engine and fuse the cover image with the cover text on the image side in an ALBEF \cite{li2021align} manner. However, the cost of image-text similarity inference becomes expensive. In addition, the image-text contrastive learning becomes highly dependent on the accuracy of the OCR engine, which creates an obstacle for model generalization.

Differently, we are inspired by \cite{kim2022ocr,davis2022end} to design UniCLIP in an OCR-free form. One idea is to guide the ViT to perceive the cover texts during the training process through agent tasks, so that it relies on no module related to the OCR function in the inference process. Since the presence of cover text is uncertain, we propose the presence-guided encoder, where the first agent task of UniCLIP is set as an Image Classification (IC) task: "To determine whether an image carries cover texts".

Specifically, as shown in Fig. \ref{fig:model}, the presence-guided encoder takes the output tokens from the last layer of ViT as input. These tokens go through a 3-layer, 8-header Transformer \cite{vaswani2017attention} structure, after which they are fed into a MLP layer for predicting the presence or absence of cover texts. The loss in this part is $L_{IC}$.

\subsection{Semantic-guided Encoder}

The presence-guided encoder directs the cover image encoder to focus on the cover texts, but does not involve semantic information. We further propose the semantic-guided encoder, which sets the second agent task as an Image-Text Matching (ITM) task: "To determine whether the specified text is consistent with the text on the cover image", encouraging the ViT to incorporate gains from the semantics of the cover texts. This design is motivated by the notion that visual tokens from the ViT contain the semantic information of cover texts, if they are successfully employed to discriminate the consistency of the cover text with a given text.

We design negative samples by nearest neighbor lookup. Take the training on CBVS-5M dataset as an example. First, we adopt RoBERTa-wwm-Base as the encoder and load the checkpoints released by QA-CLIP to encode all the 2.0M valid OCR texts. The Hierarchical Navigable Small World Algorithm (HNSW) \cite{malkov2018efficient} is applied to retrieve the Top-$K$ OCR texts that are most semantically similar but not identical to the anchor, and one of them is randomly selected as the negative sample. For covers without OCR texts, only negative samples are considered. For covers with OCR texts, the percentage of positive samples is set to 70\%.

The semantic-guided encoder accepts two inputs, i.e., tokens from the last layer of the ViT, and the embeddings of positive or negative samples. As shown in Fig. \ref{fig:model}, the module is a 3-layer structure, where each layer consists of a self-attention, a cross-attention with an MLP, and includes residual connections. The embeddings of the samples are updated layer by layer. The loss of this process is $L_{ITM}$.

\subsection{Training and Inference}
\label{sec:train}

Pre-training for UniCLIP starts with the checkpoints released by QA-CLIP. Fine-tuning is then performed on the CBVS-5M/10M dataset with a total loss of:
\begin{equation}
    L_{total}=\lambda_{1} L_{ITC}+\lambda_{2} L_{IC} + \lambda_{3} L_{ITM},
\end{equation}
where $\lambda_{1}$, $\lambda_{2}$ and $\lambda_{3}$ are hyperparameters.

It is worth noting that the fine-tuning of UniCLIP relies on positive samples, negative samples and ground truths related to OCR texts, but the inference process does not rely on any OCR-related components. As shown in Fig. \ref{fig:model}, the presence-guided encoder and semantic-guided encoder guide the training of the image-text alignment task in UniCLIP, but do not participate in the inference process. Therefore, UniCLIP infers in a manner consistent with CLIP.

\section{Experiments}

\begin{table*}
\begin{center}
  \caption{Evaluation on the CBVS-20K dataset. Our proposal achieves SOTA performance.}
  \label{tab:cmp_method}
  \setlength{\tabcolsep}{3pt} \renewcommand{\arraystretch}{1.1}
  \begin{tabular}{c|c|c|c|c|c|c|c|c|c|c}
    \toprule
    \multirow{2}{*}{Mode}&\multirow{2}{*}{Method}&\multicolumn{4}{c|}{Recall Metrics}&\multicolumn{5}{c}{Rank Metrics}\\
    \cline{3-11}
    &&R@1&R@5&R@10&MR&PNR&NDCG@1&NDCG@5&NDCG@10&MAP\\
    \midrule
    \multirow{11}{*}{Zero-shot}&CN-CLIP$_{ViT-B/16}$&0.384&0.628&0.704&0.572&2.718&0.768&0.835&0.885&0.764\\
    &CN-CLIP$_{ViT-L/14}$&0.434&0.685&0.756&0.625&2.812&0.773&0.840&0.889&0.775 \\
    &WuKong$_{ViT-B/32}$&0.197&0.356&0.439&0.331&2.000&0.702&0.791&0.858&0.712 \\
    &WuKong$_{ViT-L/14}$&0.311&0.503&0.583&0.466&2.229&0.739&0.811&0.872&0.738\\
    &Taiyi-CLIP$_{ViT-B}$&0.251&0.445&0.525&0.407&2.142&0.718&0.800&0.861&0.727\\
    &Taiyi-CLIP$_{ViT-L}$&0.269&0.492&0.577&0.446&2.278&0.726&0.805&0.866&0.733\\
    &Ernie-ViL2.0$_{ViT-B}$&0.413&0.660&0.742&0.605&2.759&0.764&0.835&0.886&0.768\\
    &R2D2-23M$_{ViT-L/14}$&0.258&0.407&0.436&0.367&2.312&0.733&0.810&0.868&0.738\\
    &R2D2-250M$_{ViT-L/14}$&0.356&0.512&0.535&0.468&2.829&\textbf{0.789}&0.842&0.891&0.775\\
    &AltCLIP$_{ViT-L}$&0.162&0.284&0.336&0.261&1.869&0.669&0.771&0.842&0.701\\
    &QA-CLIP$_{ViT-B/16}$&0.400&0.652&0.724&0.592&2.804&0.774&0.838&0.888&0.770\\
    \midrule
    \multirow{5}{*}{Fine-tuning}&CN-CLIP$_{ViT-B/16}$&0.471&0.721&0.796&0.663&2.914&0.767&0.838&0.888&0.767\\
    &R2D2-250M$_{ViT-L/14}$&0.418&0.605&0.650&0.558&2.934&0.780&0.844&0.891&0.774\\
    &QA-CLIP$_{ViT-B/16}$&0.473&0.711&0.783&0.656&2.907&0.778&0.841&0.890&0.771\\
    &ALBEF-CLIP$_{ViT-B/16}$&0.468&0.731&0.794&0.664&2.906&0.771&0.839&0.889&0.769\\
    &UniCLIP$_{ViT-B/16}$&\textbf{0.503}&\textbf{0.754}&\textbf{0.820}&\textbf{0.692}&\textbf{3.069}&0.784&\textbf{0.846}&\textbf{0.893}&\textbf{0.779}\\
  \bottomrule
\end{tabular}
\end{center}
\end{table*}

\subsection{Evaluation Metrics}

\subsubsection{Recall Metrics}

Recall is a widely adopted retrieval metric. We report on the Recall (R) and Mean Recall (MR) of the models.

\subsubsection{Rank Metrics}
Positive-to-Negative Ratio (PNR) measures the consistency of predicted results with ground truth. Formally, PNR is defined as:

\begin{equation}
\label{eq:PNR}
    PNR=\frac{\sum_{k}\sum_{i,j\in S_k}\mathbb{I}\{y_{ki}>y_{kj}\} \cdot \mathbb{I}\{\hat{y}_{ki}>\hat{y}_{kj}\}}{\sum_{k}\sum_{i,j\in S_k}\mathbb{I}\{y_{ki}>y_{kj}\} \cdot \mathbb{I}\{\hat{y}_{ki}<\hat{y}_{kj}\}},
\end{equation}
where $\mathbb{I}$ is the indicator function. The result of the indicator function is $1$ if the internal expression is true, and $0$ otherwise. $S_k$ represents the set of all documents under query $k$. $y_{ki}$, $\hat{y}_{ki}$ are the true and predicted labels of $<$image $i$, text $k$$>$, respectively. In particular, we compute PNR only for documents under the same user query.

% Normalized Discounted Cumulative Gain (NDCG) is a widely adopted metric in the field of search ranking that encourages higher rankings for better matching documents. Formally, DCG is defined as:

% \begin{equation}
%     DCG@k=\sum_{pos=1}^{k}\frac{2^{y_{pos}}-1}{log(1+pos)},
% \end{equation}
% where $y_{pos}$ is the ground truth label of the document at position $pos$. Further, IDCG is the DCG value of the ideal sort. NDCG is obtained by dividing by IDCG \cite{manning2009introduction}:
% \begin{equation}
%     NDCG@k = \frac{DCG@k}{IDCG@k}.
% \end{equation}

In addition to PNR, we also report on the Mean Average Precision (MAP) and Normalized Discounted Cumulative Gain (NDCG) metrics to fully evaluate our model.

\subsection{Implementation Details}

We follow the model architecture setup in OpenAI CLIP with ViT-B/16 as the visual backbone network for UniCLIP. The text encoder is the 12-layer architecture of RoBERTa-wwm-Base. Both are implemented by 12-layer 12-head Transformers with 768 encoding dimensions and eventually mapped linearly to 512 dimensions. The vocabulary size of text tokenizer is consistent with CN-CLIP. The initialized weights of the visual encoder and text encoder are from QA-CLIP as described in Sec. \ref{sec:train}. In addition, the weights of presence-guided encoder and semantic-guided encoder are randomly initialized with normal distribution, and $K$ for the semantic-guided encoder is set to 10.

In the fine-tuning stage, we perform random-size cropping and AutoAugment \cite{cubuk2019autoaugment} on the input image. All parameters of the image encoder and text encoder are allowed to be updated. Training is executed on 8 NVIDIA A100 GPUs for 20 epochs, with a learning rate of 2e-5. The maximum length of the text encoder is 12 and the weight decay is 1e-3. We leverage all-gather communications across GPU workers to compute $L_{ITC}$ on the global batch. The batch size is set to 1,520 due to GPU memory limitations. Mixed-precision training is activated. We save checkpoints for each epoch and report results with the highest Positive-to-Negative Ratio (PNR) on CBVS-20K. Due to computational resource constraints, we report the results of training on the CBVS-5M dataset. We simultaneously release the CBVS-10M dataset for subsequent studies.

\subsection{Comparisons}

To demonstrate the performance of our proposal and the value of the CBVS dataset, we extensively evaluate advanced Chinese image-text models on CBVS-20K. The competing models include CN-CLIP \cite{yang2022chinese}, R2D2 \cite{xie2023ccmb}, Wukong \cite{gu2022wukong}, TaiyiCLIP \cite{zhang2022fengshenbang}, Ernie-ViL2.0 \cite{shan2022ernie} and AltCLIP \cite{chen2022altclip}. The results are shown in Tab. \ref{tab:cmp_method}.

The results show that, for example, WuKong, although it achieves competitive performance on other public datasets, it lacks the ability of image-text matching on CBVS-20K. The same conclusion appears in most of the other competitors, e.g., the MR of Taiyi-CLIP$_{ViT-B}$ is only 0.407, which much lower than UniCLIP's 0.692. The generalization performance of models trained on large-scale open domain images is generally low on the CBVS-20K, which demonstrates the domain uniqueness of video cover images. Vision-Language Models that perform well in the open domain do not migrate to the video cover domain as expected.

It is worth noting that although R2D2-250M$_{ViT-L}$'s recall metrics are significantly behind UniCLIP, its ranking metrics are close to those of UniCLIP. In particular, NDCG@1 slightly outperforms UniCLIP with a maximum of 0.789. We infer that the experimental result is due to the fact that R2D2-250M$_{ViT-L}$ employs a more powerful visual architecture and enjoys a training corpus size of 250M. We encourage the incorporation of CBVS-10M into the training corpus on the one hand, and the adoption of ViT-L as a visual skeleton on the other hand, to facilitate further improvement of UniCLIP performance in subsequent studies.

Compared to the pre-trained QA-CLIP, the fine-tuning on the CBVS-5M dataset comprehensively improves the metrics, especially the PNR by 3.67\%, and R@1 by 18.25\%. Performing fine-tuning on the publicly released CN-CLIP$_{ViT-B}$, consistent findings are observed with a 7.21\% improvement in PNR and a 22.66\% improvement in R@1. Performing fine-tuning on the R2D2-250M$_{ViT-L}$, significantly higher recall metrics are observed, as well as largely comparable rank metrics. These results demonstrate that fine-tuning on a large-scale cover dataset can improve the performance of the model in the video search domain. In addition, UniCLIP achieves state-of-the-art performance with the highest metrics compared to its competitors and does not introduce additional inference cost compared to the simple and efficient CN-CLIP.

\begin{table*}
\begin{center}
    \caption{Results of ablation study of UniCLIP.}
    \label{tab:ablation}
    \setlength{\tabcolsep}{3pt} \renewcommand{\arraystretch}{1.1}
    \begin{tabular}{c|c|c|c|c|c|c|c|c|c|c}
    \toprule
    \multirow{2}{*}{$L_{IC}$} & \multirow{2}{*}{$L_{ITM}$}&\multicolumn{4}{c|}{Recall Metrics}&\multicolumn{5}{c}{Rank Metrics}\\
    \cline{3-11}
    &&R@1&R@5&R@10&MR&PNR&NDCG@1&NDCG@5&NDCG@10&MAP\\
    \midrule
    &&0.473&0.711&0.783&0.656&2.907&0.778&0.841&0.890&0.771\\
    \checkmark & &0.491&0.747&0.818&0.685&2.991&0.776&0.843&0.890&0.772\\
    & \checkmark &0.499&\textbf{0.754}&0.812&0.688&3.006&0.783&0.845&\textbf{0.893}&\textbf{0.779}\\
    \checkmark & \checkmark &\textbf{0.503}&\textbf{0.754}&\textbf{0.820}&\textbf{0.692}&\textbf{3.069}&\textbf{0.784}&\textbf{0.846}&\textbf{0.893}&\textbf{0.779}\\
    \bottomrule
    \end{tabular}
\end{center}
\end{table*}

\begin{table*}
\begin{center}
    \caption{PNR metrics for different OCR texts combinations. $<S_T, S_F>$ stands for that in Eq. \ref{eq:PNR}, $i \in S_T$ and $j \in S_F$.}
    \label{tab:ocr}
    \begin{tabular}{c|c|c|c|c}
    \toprule
    \multirow{2}{*}{Model}&$<S_{T}, S_{T}>$&$<S_{F}, S_{F}>$&$<S_{T}, S_{F}>$&All\\
    &(11.71\%)&(46.51\%)&(41.78\%)&(100.00\%)\\
    \midrule
    QA-CLIP$_{ViT-B/16}$&3.203&2.722&2.975&2.877\\
    ALBEF-CLIP$_{ViT-B/16}$&\textbf{3.375}&2.689&3.051&2.906\\
    UniCLIP$_{ViT-B/16}$&3.331&\textbf{2.904}&\textbf{3.194}&\textbf{3.069}\\
    \bottomrule
    \end{tabular}
\end{center}
\end{table*}

\subsection{Ablation Study}

% Compared to the zero-shot QA-CLIP, the gain of UniCLIP comes from 3 aspects: introduction of OCR modalities during training, presence-guided encoder and semantic-guided encoder. Tab. \ref{tab:ablation} shows the results of the ablation study of UniCLIP.

In order to evaluate the proposed presence-guided encoder and semantic-guided encoder, we implement versions of the model with and without $L_{IC}$ and $L_{ITM}$, respectively. The weights of $L_{ITC}$ and $L_{IC}$ (or $L_{ITM}$) are set to 0.8 and 0.2. Tab. \ref{tab:ablation} shows the results of the ablation study of UniCLIP.

% If both the presence-guided encoder and the semantic-guided encoder are removed and the cover text is discarded, the model degenerates into a fine-tuned version of QA-CLIP. The PNR is reduced from 3.069 to 2.907 and the MR from 0.692 to 0.656. This result demonstrates the information gain of the cover text.

If both the presence-guided encoder and the semantic-guided encoder are removed and the cover text is discarded, the model degenerates into a fine-tuned version of QA-CLIP. The PNR is reduced from 3.069 to 2.907 and the MR from 0.692 to 0.656. In addition, removing any of the two encoders results in varying degrees of degradation in model performance. Removing the presence-guided encoder reduces the PNR of UniCLIP by 2.05\%. Besides, removing the semantic-guided encoder reduces the PNR by 2.54\%. Interestingly, when only the semantic-guided encoder is employed, the NDCG@1/5/10, MAP, and R@1/5 of the model are basically the same as the final scheme, which indicates that the gain of cover text is mainly in the semantic information. If two encoders are employed at the same time, i.e., the proposed two agent tasks are considered at the same time, the highest metrics are achieved in all aspects. This result is ample proof of the validity of our proposal. We encourage follow-up studies to further generalise our ideas.

% \begin{figure}
% \begin{center}
% \includegraphics[width=\linewidth]{imgs/albef.pdf}
% \end{center}
%    \caption{An explicit OCR fusion scheme.}
% \label{fig:ocr}
% \end{figure}

\subsection{Cover Text Capability Assessment}

Since the training process of UniCLIP relies on the cover text modality, for a fair comparison, we implement an explicit OCR text fusion scheme, which is denoted as ALBEF-CLIP. Compared to CLIP, we replace ViT with an ALBEF structure, where the cover image and the cover text go through their respective encoders before passing through a 6-layer Attention-based fusion structure. For the case of missing OCR texts, the prompt word is employed. The results of ALBEF-CLIP are shown in Tab. \ref{tab:cmp_method}. If the cover text is introduced, but with ALBEF-CLIP rather than our proposal to exploit this modality, the metrics are much lower than UniCLIP in all aspects. We hypothesize that the reason for this is that UniCLIP guides the semantic training of ViT and handles the modality missing problem more consistently, reducing information confusion.

To further evaluate the cover text capability, we categorize the data in CBVS-20K into two main categories according to the presence or absence of cover texts, which are denoted as $S_T$ and $S_F$, respectively. Tab. \ref{tab:ocr} demonstrates the PNR metrics for different combinations. Compared to the scheme without cover texts (QA-CLIP), ALBEF-CLIP significantly improves the matching ability for covers with cover texts, increasing the PNR from 3.203 to 3.375. However, for covers without cover texts, the scheme degrades the performance, which may be due to semantic confusions brought about by prompt words.

In comparison, UniCLIP is basically comparable to ALBEF-CLIP for matching between covers with cover texts. This is in line with expectations, as we discard cover texts modalities in our inference, however the very close results are a good indication of the promise of UniCLIP. Besides, UniCLIP performs much better for both other cases than QA-CLIP and ALBEF-CLIP. For the matching between covers without cover texts, which is most likely to happen, UniCLIP's PNR exceeds that of the fusion scheme by 8.00\% and has a lower inference cost. For the hybrid cases, UniCLIP achieves the PNR of 3.194. This suggests that UniCLIP is able to overcome the modality missing problem to some extent and handle cover images with or without cover texts uniformly. Thanks to this, UniCLIP shows the best performance on the full range of data.

% To further evaluate the OCR capability of UniCLIP, we categorize the data in CBVS-20K into two main categories according to the presence or absence of OCR, which are denoted as $S_T$ and $S_F$, respectively. Tab. \ref{tab:ocr} demonstrates the PNR metrics for different combinations. Compared to the scheme without OCR, the OCR fusion scheme improves the matching ability for covers with OCR, increasing the PNR from 3.203 to 3.375. However, for covers without OCR, the scheme degrades the performance, which may be due to semantic confusions brought about by prompt words. In comparison, UniCLIP is basically comparable to the fusion scheme for matching between covers with OCR, but performs much better for both other cases. This is in line with expectations, as we discard OCR modalities in our inference. These results suggest that UniCLIP is able to overcome the modality missing problem to some extent and handle cover images with or without OCR uniformly.

% \begin{figure*}
% \begin{center}
% \includegraphics[width=0.9\linewidth]{imgs/categories.png}
% \end{center}
%    \caption{zx.}
% \label{fig:category}
% \end{figure*}

\section{Conclusion}

In this work, we release the largest publicly available Chinese video cover-video title dataset to fill in the lack of cover data for short video search scenarios. We further build a manual fine-labeling video cover-user query benchmark test for short video search domain and propose UniCLIP to unify cover texts to guide contrastive learning. The image classification task and the image-text matching task are performed in an OCR-free manner. We believe in the ability of CBVS-5M/10M to expand the domain of large-scale Chinese image-text training and are pleasantly surprised to observe the substrate-independent potential of UniCLIP. However, significant room for exploration remains in balancing the video search domain with the generalized domain. We look forward to the extension of CBVS to downstream tasks such as title generation, as well as inspiration from UniCLIP for performing multi-modal fusion in the CLIP framework.

\bibliographystyle{named}
\bibliography{ijcai24}

\begin{thebibliography}{}

\bibitem[\protect\citeauthoryear{Abdullah and Rangarajan}{2021}]{abdullah2021image}
Taghreed Abdullah and Lalitha Rangarajan.
\newblock Image-text matching: Methods and challenges.
\newblock {\em Inventive Systems and Control: Proceedings of ICISC 2021}, pages 213--222, 2021.

\bibitem[\protect\citeauthoryear{Chen \bgroup \em et al.\egroup }{2015}]{chen2015microsoft}
Xinlei Chen, Hao Fang, Tsung-Yi Lin, Ramakrishna Vedantam, Saurabh Gupta, Piotr Doll{\'a}r, and C~Lawrence Zitnick.
\newblock Microsoft coco captions: Data collection and evaluation server. arxiv 2015.
\newblock {\em arXiv preprint arXiv:1504.00325}, 2015.

\bibitem[\protect\citeauthoryear{Chen \bgroup \em et al.\egroup }{2020}]{chen2020imram}
Hui Chen, Guiguang Ding, Xudong Liu, Zijia Lin, Ji~Liu, and Jungong Han.
\newblock Imram: Iterative matching with recurrent attention memory for cross-modal image-text retrieval.
\newblock In {\em Proceedings of the IEEE/CVF conference on computer vision and pattern recognition}, pages 12655--12663, 2020.

\bibitem[\protect\citeauthoryear{Chen \bgroup \em et al.\egroup }{2021}]{chen2021learning}
Jiacheng Chen, Hexiang Hu, Hao Wu, Yuning Jiang, and Changhu Wang.
\newblock Learning the best pooling strategy for visual semantic embedding.
\newblock In {\em Proceedings of the IEEE/CVF conference on computer vision and pattern recognition}, pages 15789--15798, 2021.

\bibitem[\protect\citeauthoryear{Chen \bgroup \em et al.\egroup }{2022}]{chen2022altclip}
Zhongzhi Chen, Guang Liu, Bo-Wen Zhang, Fulong Ye, Qinghong Yang, and Ledell Wu.
\newblock Altclip: Altering the language encoder in clip for extended language capabilities.
\newblock {\em arXiv preprint arXiv:2211.06679}, 2022.

\bibitem[\protect\citeauthoryear{Cubuk \bgroup \em et al.\egroup }{2019}]{cubuk2019autoaugment}
Ekin~D Cubuk, Barret Zoph, Dandelion Mane, Vijay Vasudevan, and Quoc~V Le.
\newblock Autoaugment: Learning augmentation strategies from data.
\newblock In {\em Proceedings of the IEEE/CVF conference on computer vision and pattern recognition}, pages 113--123, 2019.

\bibitem[\protect\citeauthoryear{Davis \bgroup \em et al.\egroup }{2022}]{davis2022end}
Brian Davis, Bryan Morse, Brian Price, Chris Tensmeyer, Curtis Wigington, and Vlad Morariu.
\newblock End-to-end document recognition and understanding with dessurt.
\newblock In {\em European Conference on Computer Vision}, pages 280--296. Springer, 2022.

\bibitem[\protect\citeauthoryear{Devlin \bgroup \em et al.\egroup }{2018}]{devlin2018bert}
Jacob Devlin, Ming-Wei Chang, Kenton Lee, and Kristina Toutanova.
\newblock Bert: Pre-training of deep bidirectional transformers for language understanding.
\newblock {\em arXiv preprint arXiv:1810.04805}, 2018.

\bibitem[\protect\citeauthoryear{Diao \bgroup \em et al.\egroup }{2021}]{diao2021similarity}
Haiwen Diao, Ying Zhang, Lin Ma, and Huchuan Lu.
\newblock Similarity reasoning and filtration for image-text matching.
\newblock In {\em Proceedings of the AAAI conference on artificial intelligence}, volume~35, pages 1218--1226, 2021.

\bibitem[\protect\citeauthoryear{Dosovitskiy \bgroup \em et al.\egroup }{2020}]{dosovitskiy2020image}
Alexey Dosovitskiy, Lucas Beyer, Alexander Kolesnikov, Dirk Weissenborn, Xiaohua Zhai, Thomas Unterthiner, Mostafa Dehghani, Matthias Minderer, Georg Heigold, Sylvain Gelly, et~al.
\newblock An image is worth 16x16 words: Transformers for image recognition at scale.
\newblock {\em arXiv preprint arXiv:2010.11929}, 2020.

\bibitem[\protect\citeauthoryear{Faghri \bgroup \em et al.\egroup }{2017}]{faghri2017vse++}
Fartash Faghri, David~J Fleet, Jamie~Ryan Kiros, and Sanja Fidler.
\newblock Vse++: Improving visual-semantic embeddings with hard negatives.
\newblock {\em arXiv preprint arXiv:1707.05612}, 2017.

\bibitem[\protect\citeauthoryear{Fu \bgroup \em et al.\egroup }{2023}]{fu2023learning}
Zheren Fu, Zhendong Mao, Yan Song, and Yongdong Zhang.
\newblock Learning semantic relationship among instances for image-text matching.
\newblock In {\em Proceedings of the IEEE/CVF Conference on Computer Vision and Pattern Recognition}, pages 15159--15168, 2023.

\bibitem[\protect\citeauthoryear{Gan \bgroup \em et al.\egroup }{2023}]{gan2023cnvid}
Tian Gan, Qing Wang, Xingning Dong, Xiangyuan Ren, Liqiang Nie, and Qingpei Guo.
\newblock Cnvid-3.5 m: Build, filter, and pre-train the large-scale public chinese video-text dataset.
\newblock In {\em Proceedings of the IEEE/CVF Conference on Computer Vision and Pattern Recognition}, pages 14815--14824, 2023.

\bibitem[\protect\citeauthoryear{Gu \bgroup \em et al.\egroup }{2022}]{gu2022wukong}
Jiaxi Gu, Xiaojun Meng, Guansong Lu, Lu~Hou, Niu Minzhe, Xiaodan Liang, Lewei Yao, Runhui Huang, Wei Zhang, Xin Jiang, et~al.
\newblock Wukong: A 100 million large-scale chinese cross-modal pre-training benchmark.
\newblock {\em Advances in Neural Information Processing Systems}, 35:26418--26431, 2022.

\bibitem[\protect\citeauthoryear{Hendriksen \bgroup \em et al.\egroup }{2022}]{hendriksen2022extending}
Mariya Hendriksen, Maurits Bleeker, Svitlana Vakulenko, Nanne van Noord, Ernst Kuiper, and Maarten de~Rijke.
\newblock Extending clip for category-to-image retrieval in e-commerce.
\newblock In {\em European Conference on Information Retrieval}, pages 289--303. Springer, 2022.

\bibitem[\protect\citeauthoryear{Kim \bgroup \em et al.\egroup }{2022}]{kim2022ocr}
Geewook Kim, Teakgyu Hong, Moonbin Yim, JeongYeon Nam, Jinyoung Park, Jinyeong Yim, Wonseok Hwang, Sangdoo Yun, Dongyoon Han, and Seunghyun Park.
\newblock Ocr-free document understanding transformer.
\newblock In {\em European Conference on Computer Vision}, pages 498--517. Springer, 2022.

\bibitem[\protect\citeauthoryear{Krishna \bgroup \em et al.\egroup }{2017}]{krishna2017visual}
Ranjay Krishna, Yuke Zhu, Oliver Groth, Justin Johnson, Kenji Hata, Joshua Kravitz, Stephanie Chen, Yannis Kalantidis, Li-Jia Li, David~A Shamma, et~al.
\newblock Visual genome: Connecting language and vision using crowdsourced dense image annotations.
\newblock {\em International journal of computer vision}, 123:32--73, 2017.

\bibitem[\protect\citeauthoryear{Lei \bgroup \em et al.\egroup }{2021}]{lei2021understanding}
Chenyi Lei, Shixian Luo, Yong Liu, Wanggui He, Jiamang Wang, Guoxin Wang, Haihong Tang, Chunyan Miao, and Houqiang Li.
\newblock Understanding chinese video and language via contrastive multimodal pre-training.
\newblock In {\em Proceedings of the 29th ACM International Conference on Multimedia}, pages 2567--2576, 2021.

\bibitem[\protect\citeauthoryear{Li \bgroup \em et al.\egroup }{2021}]{li2021align}
Junnan Li, Ramprasaath Selvaraju, Akhilesh Gotmare, Shafiq Joty, Caiming Xiong, and Steven Chu~Hong Hoi.
\newblock Align before fuse: Vision and language representation learning with momentum distillation.
\newblock {\em Advances in neural information processing systems}, 34:9694--9705, 2021.

\bibitem[\protect\citeauthoryear{Li \bgroup \em et al.\egroup }{2022}]{li2022grounded}
Liunian~Harold Li, Pengchuan Zhang, Haotian Zhang, Jianwei Yang, Chunyuan Li, Yiwu Zhong, Lijuan Wang, Lu~Yuan, Lei Zhang, Jenq-Neng Hwang, et~al.
\newblock Grounded language-image pre-training.
\newblock In {\em Proceedings of the IEEE/CVF Conference on Computer Vision and Pattern Recognition}, pages 10965--10975, 2022.

\bibitem[\protect\citeauthoryear{Lin \bgroup \em et al.\egroup }{2021}]{lin2021m6}
Junyang Lin, Rui Men, An~Yang, Chang Zhou, Ming Ding, Yichang Zhang, Peng Wang, Ang Wang, Le~Jiang, Xianyan Jia, et~al.
\newblock M6: A chinese multimodal pretrainer.
\newblock {\em arXiv preprint arXiv:2103.00823}, 2021.

\bibitem[\protect\citeauthoryear{Liu \bgroup \em et al.\egroup }{2019}]{liu2019roberta}
Yinhan Liu, Myle Ott, Naman Goyal, Jingfei Du, Mandar Joshi, Danqi Chen, Omer Levy, Mike Lewis, Luke Zettlemoyer, and Veselin Stoyanov.
\newblock Roberta: A robustly optimized bert pretraining approach.
\newblock {\em arXiv preprint arXiv:1907.11692}, 2019.

\bibitem[\protect\citeauthoryear{Liu \bgroup \em et al.\egroup }{2020}]{liu2020graph}
Chunxiao Liu, Zhendong Mao, Tianzhu Zhang, Hongtao Xie, Bin Wang, and Yongdong Zhang.
\newblock Graph structured network for image-text matching.
\newblock In {\em Proceedings of the IEEE/CVF conference on computer vision and pattern recognition}, pages 10921--10930, 2020.

\bibitem[\protect\citeauthoryear{Malkov and Yashunin}{2018}]{malkov2018efficient}
Yu~A Malkov and Dmitry~A Yashunin.
\newblock Efficient and robust approximate nearest neighbor search using hierarchical navigable small world graphs.
\newblock {\em IEEE transactions on pattern analysis and machine intelligence}, 42(4):824--836, 2018.

\bibitem[\protect\citeauthoryear{Mu \bgroup \em et al.\egroup }{2022}]{mu2022slip}
Norman Mu, Alexander Kirillov, David Wagner, and Saining Xie.
\newblock Slip: Self-supervision meets language-image pre-training.
\newblock In {\em European Conference on Computer Vision}, pages 529--544. Springer, 2022.

\bibitem[\protect\citeauthoryear{Nie \bgroup \em et al.\egroup }{2022}]{nie2022search}
Liqiang Nie, Leigang Qu, Dai Meng, Min Zhang, Qi~Tian, and Alberto~Del Bimbo.
\newblock Search-oriented micro-video captioning.
\newblock In {\em Proceedings of the 30th ACM International Conference on Multimedia}, pages 3234--3243, 2022.

\bibitem[\protect\citeauthoryear{Oord \bgroup \em et al.\egroup }{2018}]{oord2018representation}
Aaron van~den Oord, Yazhe Li, and Oriol Vinyals.
\newblock Representation learning with contrastive predictive coding.
\newblock {\em arXiv preprint arXiv:1807.03748}, 2018.

\bibitem[\protect\citeauthoryear{Qu \bgroup \em et al.\egroup }{2020}]{qu2020context}
Leigang Qu, Meng Liu, Da~Cao, Liqiang Nie, and Qi~Tian.
\newblock Context-aware multi-view summarization network for image-text matching.
\newblock In {\em Proceedings of the 28th ACM International Conference on Multimedia}, pages 1047--1055, 2020.

\bibitem[\protect\citeauthoryear{Radford \bgroup \em et al.\egroup }{2021}]{radford2021learning}
Alec Radford, Jong~Wook Kim, Chris Hallacy, Aditya Ramesh, Gabriel Goh, Sandhini Agarwal, Girish Sastry, Amanda Askell, Pamela Mishkin, Jack Clark, et~al.
\newblock Learning transferable visual models from natural language supervision.
\newblock In {\em International conference on machine learning}, pages 8748--8763. PMLR, 2021.

\bibitem[\protect\citeauthoryear{Shan \bgroup \em et al.\egroup }{2022}]{shan2022ernie}
Bin Shan, Weichong Yin, Yu~Sun, Hao Tian, Hua Wu, and Haifeng Wang.
\newblock Ernie-vil 2.0: Multi-view contrastive learning for image-text pre-training.
\newblock {\em arXiv preprint arXiv:2209.15270}, 2022.

\bibitem[\protect\citeauthoryear{Spola{\^o}r \bgroup \em et al.\egroup }{2020}]{spolaor2020systematic}
Newton Spola{\^o}r, Huei~Diana Lee, Weber Shoity~Resende Takaki, Leandro~Augusto Ensina, Claudio Saddy~Rodrigues Coy, and Feng~Chung Wu.
\newblock A systematic review on content-based video retrieval.
\newblock {\em Engineering Applications of Artificial Intelligence}, 90:103557, 2020.

\bibitem[\protect\citeauthoryear{Vaswani \bgroup \em et al.\egroup }{2017}]{vaswani2017attention}
Ashish Vaswani, Noam Shazeer, Niki Parmar, Jakob Uszkoreit, Llion Jones, Aidan~N Gomez, {\L}ukasz Kaiser, and Illia Polosukhin.
\newblock Attention is all you need.
\newblock {\em Advances in neural information processing systems}, 30, 2017.

\bibitem[\protect\citeauthoryear{Wang \bgroup \em et al.\egroup }{2019}]{wang2019vatex}
Xin Wang, Jiawei Wu, Junkun Chen, Lei Li, Yuan-Fang Wang, and William~Yang Wang.
\newblock Vatex: A large-scale, high-quality multilingual dataset for video-and-language research.
\newblock In {\em Proceedings of the IEEE/CVF International Conference on Computer Vision}, pages 4581--4591, 2019.

\bibitem[\protect\citeauthoryear{Wray \bgroup \em et al.\egroup }{2021}]{wray2021semantic}
Michael Wray, Hazel Doughty, and Dima Damen.
\newblock On semantic similarity in video retrieval.
\newblock In {\em Proceedings of the IEEE/CVF Conference on Computer Vision and Pattern Recognition}, pages 3650--3660, 2021.

\bibitem[\protect\citeauthoryear{Xie \bgroup \em et al.\egroup }{2023}]{xie2023ccmb}
Chunyu Xie, Heng Cai, Jincheng Li, Fanjing Kong, Xiaoyu Wu, Jianfei Song, Henrique Morimitsu, Lin Yao, Dexin Wang, Xiangzheng Zhang, et~al.
\newblock Ccmb: A large-scale chinese cross-modal benchmark.
\newblock In {\em Proceedings of the 31st ACM International Conference on Multimedia}, pages 4219--4227, 2023.

\bibitem[\protect\citeauthoryear{Xu \bgroup \em et al.\egroup }{2023}]{xu2023youku}
Haiyang Xu, Qinghao Ye, Xuan Wu, Ming Yan, Yuan Miao, Jiabo Ye, Guohai Xu, Anwen Hu, Yaya Shi, Guangwei Xu, et~al.
\newblock Youku-mplug: A 10 million large-scale chinese video-language dataset for pre-training and benchmarks.
\newblock {\em arXiv preprint arXiv:2306.04362}, 2023.

\bibitem[\protect\citeauthoryear{Yang \bgroup \em et al.\egroup }{2022}]{yang2022chinese}
An~Yang, Junshu Pan, Junyang Lin, Rui Men, Yichang Zhang, Jingren Zhou, and Chang Zhou.
\newblock Chinese clip: Contrastive vision-language pretraining in chinese.
\newblock {\em arXiv preprint arXiv:2211.01335}, 2022.

\bibitem[\protect\citeauthoryear{Zhan \bgroup \em et al.\egroup }{2021}]{zhan2021product1m}
Xunlin Zhan, Yangxin Wu, Xiao Dong, Yunchao Wei, Minlong Lu, Yichi Zhang, Hang Xu, and Xiaodan Liang.
\newblock Product1m: Towards weakly supervised instance-level product retrieval via cross-modal pretraining.
\newblock In {\em Proceedings of the IEEE/CVF International Conference on Computer Vision}, pages 11782--11791, 2021.

\bibitem[\protect\citeauthoryear{Zhang \bgroup \em et al.\egroup }{2020}]{zhang2020poet}
Shengyu Zhang, Ziqi Tan, Jin Yu, Zhou Zhao, Kun Kuang, Jie Liu, Jingren Zhou, Hongxia Yang, and Fei Wu.
\newblock Poet: Product-oriented video captioner for e-commerce.
\newblock In {\em Proceedings of the 28th ACM International Conference on Multimedia}, pages 1292--1301, 2020.

\bibitem[\protect\citeauthoryear{Zhang \bgroup \em et al.\egroup }{2021}]{zhang2021tip}
Renrui Zhang, Rongyao Fang, Wei Zhang, Peng Gao, Kunchang Li, Jifeng Dai, Yu~Qiao, and Hongsheng Li.
\newblock Tip-adapter: Training-free clip-adapter for better vision-language modeling.
\newblock {\em arXiv preprint arXiv:2111.03930}, 2021.

\bibitem[\protect\citeauthoryear{Zhang \bgroup \em et al.\egroup }{2022a}]{zhang2022fengshenbang}
Jiaxing Zhang, Ruyi Gan, Junjie Wang, Yuxiang Zhang, Lin Zhang, Ping Yang, Xinyu Gao, Ziwei Wu, Xiaoqun Dong, Junqing He, et~al.
\newblock Fengshenbang 1.0: Being the foundation of chinese cognitive intelligence.
\newblock {\em arXiv preprint arXiv:2209.02970}, 2022.

\bibitem[\protect\citeauthoryear{Zhang \bgroup \em et al.\egroup }{2022b}]{zhang2022create}
Ziqi Zhang, Yuxin Chen, Zongyang Ma, Zhongang Qi, Chunfeng Yuan, Bing Li, Ying Shan, and Weiming Hu.
\newblock Create: A benchmark for chinese short video retrieval and title generation.
\newblock {\em arXiv preprint arXiv:2203.16763}, 2022.

\bibitem[\protect\citeauthoryear{Zhou \bgroup \em et al.\egroup }{2022}]{zhou2022learning}
Kaiyang Zhou, Jingkang Yang, Chen~Change Loy, and Ziwei Liu.
\newblock Learning to prompt for vision-language models.
\newblock {\em International Journal of Computer Vision}, 130(9):2337--2348, 2022.

\end{thebibliography}

\end{document}